\documentclass[fleqn,11pt]{wlscirep}
\usepackage[utf8]{inputenc}
\usepackage[T1]{fontenc}
\usepackage{setspace} 
\usepackage{lineno}

\title{Fourier neural operator for real-time simulation of 3D dynamic urban microclimate}

\author[1,3,4]{Wenhui Peng}
\author[1,2]{Shaoxiang Qin}
\author[1]{Senwen Yang}
\author[3]{Jianchun Wang}
\author[2]{Xue Liu}
\author[1,*]{Liangzhu (Leon) Wang}
\affil[1]{Concordia University, Centre for Zero Energy Building Studies, Department of Building, Civil and Environmental Engineering, Montreal, H3G 1M8, Canada}
\affil[2]{McGill University, School of Computer Science, Montreal, H3A 0G4, Canada}
\affil[3]{Southern University of Science and Technology, Department of Mechanics and Aerospace Engineering, Shenzhen, 518055, China}
\affil[4]{The Hong Kong Polytechnic University, Department  of  Applied  Mathematics, Hong Kong SAR, 999077, China}

\affil[*]{leon.wang@concordia.ca}

\begin{abstract}
Global urbanization has underscored the significance of urban microclimates for human comfort, health, and building/urban energy efficiency. They profoundly influence building design and urban planning as major environmental impacts. Understanding local microclimates is essential for cities to prepare for climate change and effectively implement resilience measures. However, analyzing urban microclimates requires considering a complex array of outdoor parameters within computational domains at the city scale over a longer period than indoors. As a result, numerical methods like Computational Fluid Dynamics (CFD) become computationally expensive when evaluating the impact of urban microclimates. The rise of deep learning techniques has opened new opportunities for accelerating the modeling of complex non-linear interactions and system dynamics. Recently, the Fourier Neural Operator (FNO) has been shown to be very promising in accelerating solving the Partial Differential Equations (PDEs) and modeling fluid dynamic systems. In this work, we apply the FNO network for real-time three-dimensional (3D) urban wind field simulation. The training and testing data are generated from CFD simulation of the urban area, based on the semi-Lagrangian approach and fractional stepping method to simulate urban microclimate features for modeling large-scale urban problems. Numerical experiments show that the FNO model can accurately reconstruct the instantaneous spatial velocity field. We further evaluate the trained FNO model on unseen data with different wind directions, and the results show that the FNO model can generalize well on different wind directions. More importantly, the FNO approach can make predictions within milliseconds on the graphics processing unit, making real-time simulation of 3D dynamic urban microclimate possible.

\end{abstract}

\begin{document}
\flushbottom
\maketitle
%
%
\thispagestyle{empty}


\section{Introduction}

Over the past decade, the world has witnessed significant urbanization and rapid industrialization, resulting in heightened vulnerability and degradation of the urban environment. The relentless expansion of urban areas involves the conversion of natural landscapes into man-made environments with unique physical characteristics. As the climate undergoes continuous change, these evolving urban features directly and profoundly impact the local living environment, giving rise to urban microclimates where the majority of human activities take place. Consequently, a surge of research has emerged, dedicated to comprehending and studying these urban microclimates. Four approaches are commonly used to study urban microclimates: field measurement and meteorological observation \cite{du2016influences,bernard2017urban}, wind tunnel experiments \cite{uehara2000wind}, computational fluid dynamics (CFD) simulation \cite{antoniou2019cfd,toparlar2015cfd,janssen2013pedestrian}, and recent advancements in artificial intelligence and machine learning techniques \cite{hab2016data}. The urban microclimate research originated in the 19th century with Howard's city heat island description, followed by the development of wind tunnel techniques in the 1990s for studying wind environment and comfort. CFD simulations gained popularity in the 1990s with the advent of commercial software tools, and their usage increased significantly in the 2010s due to improved computing power. Field measurements with higher resolution and advanced capabilities also became more prevalent. In recent years, artificial intelligence and data-driven methods have been applied to classify microclimate zones, and machine learning has been used to predict air temperature based on landscape features. Since 2010, the use of conservation equations in urban microclimate research has increased due to improved computing power for CFD. Wind tunnel experiments and field measurements are also popular approaches, but data-driven studies have seen significant growth since 2015, suggesting a changing trend in research \cite{yang2023urban}.

For predicting the urban wind environment, traditionally, field measurement and wind tunnel experiments have been employed to analyze pedestrian-level wind. However, with the advancement of computing power, CFD has become increasingly utilized. Blocken et al. \cite{blocken2016pedestrian}conducted a comprehensive review comparing wind tunnel experiments and CFD methods for pedestrian-level wind analysis. In wind tunnels, techniques like hot-wire anemometry and laser Doppler velocimetry are used to measure wind velocity and these methods provide rapid temporal response and enable measurement of both mean flow and turbulence. \textcolor{black}{Chew et al. studied the flow dynamics inside urban canyons of different aspect ratio in water channels\cite{chew2018flows}.} More recently, particle image velocimetry has been utilized for high-density spatial distribution measurements of wind velocity. Numerous studies have utilized CFD with large eddy simulation (LES) to analyze unsteady turbulent flow fields \cite{maulik2018data,wang2021artificial,park2021toward}.  \textcolor{black}{LES will become computationally costly in city-scale scenarios, while many techniques enable its use up-to scale of multiple neighborhoods. Previous studies make LES feasible at smaller neighborhood scales \cite{ mortezazadeh2019slac}. For example, the PALM package \cite{maronga2015parallelized}, the Fast Fluid Dynamics (FFD) , and Very Large Eddy Simulation (VLES) models \cite{aliabadi2018very,ahmadi2021very} over-come the computational expense to some level at smaller neighborhood scales.} In complex urban areas, turbulence models based on the Reynolds-averaged Navier-Stokes (RANS) equations are commonly employed to predict wind velocity in complex urban areas\cite{blocken2018over,zheng2020cfd,salim2011numerical,ricci2020impact}. While RANS-based models may have lower accuracy compared to LES, their faster calculation speeds make them more practical for evaluating wind velocity.

Data-driven models applied to urban physics encompass four key data sources: historical weather station records, field measurements, wind tunnel measurements, and CFD simulations. These diverse datasets serve as the foundation for research in this domain. Previous studies have diligently explored numerous parameters and extensively documented the training performance of various models. The primary focus of these investigations lies in the accurate prediction of essential variables such as wind speed, temperature, and building energy consumption. \cite{yang2023urban}. Artificial intelligence models including multiple linear regression (MLR) \cite{alonso2020new,oukawa2022fine}, nonlinear regression (NLR) \cite{alonso2020new,oukawa2022fine}, random forest (RF) \cite{alonso2020new,oukawa2022fine}, and artificial neural networks (ANN) \cite{higgins2021application,nazarian2017predicting} have been utilized based on the parameters of interest. For predicting the urban wind environment, Mortezazadeh et al. \cite{mortezazadeh2022estimating}utilized machine learning with CFD simulation results to predict wind speed and power, assessing the wind power potential in an urban area. In a recent study, Zhang et al. \cite{zhang2021urban} employed a recurrent neural network, specifically the long short-term memory (LSTM) model, to model the time-series temperature variations within an urban street canyon. They also used the LSTM model to forecast wind speed, direction, relative humidity, and solar radiation. Higgins and Stathopoulos \cite{higgins2021application} conducted a study focusing on generating an extensive database for using AI tools to improve urban wind energy generation. A decisional flow chart approach was developed, and expert and artificial neural network (ANN) systems were tested and compared based on the dataset of wind tunnel experiments and CFD simulations analyzing the impact of building shapes and urban configurations on wind velocities. Javanroodi et al. \cite{javanroodi2022combining}proposed a hybrid model combining CFD and ANN was proposed in this research to predict the interactions between climate variables and urban morphology. By using the validated CFD model and urban canopy parameterization, the vertical structure of mean wind speed and air temperature was quantified from mesoscale to microscale over and within the idealized urban areas. Microclimate and morphological databases were then utilized to train two artificial neural network models, including a Multilayer Perceptron (MLP) and a Deep Neural Network (DNN), to predict wind speed magnitudes. \textcolor{black}{Graph Neural Networks (GNNs) are particularly efficient at dealing with irregular geometry and boundary conditions \cite{li2020neural,liu2022graph,chen2021graph}, however, the computational cost directly scales with
the number of edges\cite{zhou2020graph}.}

\textcolor{black}{Physics-Informed Neural Networks (PINNs) play an important role in solving fluid dynamical systems described by PDEs. PINNs combine neural networks with principles from physics to solve problems related to fluid dynamics and other physical systems. 
In PINN, physical laws or domain expertise are incorporated into the machine learning models as a form of regularization \cite{raissi2019physics}. This structured information enhances the algorithm's ability to learn from limited data and generalize effectively.
Erichson et al. \cite{erichson2019physics} incorporated Lyapunov stability analysis into neural networks, which can enhance model quality, improve generalization performance, reduce sensitivity to parameter tuning, and enhance robustness in the presence of noisy data.
Mao et al. \cite{mao2020physics} applied PINNs to model Euler equations in high-speed aerodynamics. They explored both one-dimensional and two-dimensional domains, finding that while PINNs may be less accurate than traditional numerical methods for forward problems, they excel in solving challenging inverse problems.
Wang et al. \cite{wang2020towards} proposed a hybrid approach that combines established turbulent flow simulation techniques with deep learning, introducing trainable spectral filters and a specialized U-net model. 
The approach demonstrated significant error reductions in predicting turbulent flow characteristics.
Raissi et al. \cite{raissi2020hidden} introduced a novel approach called hidden fluid mechanics (HFM). It uses physics-informed deep learning to encode the Navier-Stokes equations into neural networks, making it applicable to various geometries and initial/boundary conditions. HFM is effective in solving physical and biomedical problems, especially in scenarios where direct measurements are challenging due to low-resolution and noisy observation data.
Xu et al. \cite{xu2021explore} highlighted the challenge of missing or unresolved data in experimental measurements of turbulent flows and proposed a solution using physics-informed deep learning. By treating the governing equations as parameterized constraints, missing flow dynamics in regions can be reconstructed without data.}

Most existing DNNs have focused on learning mappings between finite-dimensional Euclidean spaces, they are good at learning a single instance of the equation, but have difficulty to generalize well once the governing equation parameters or boundary conditions changes \cite{pan2020physics,wu2020data,xu2021deep}. Recently, Li et al. proposed the Fourier Neural Operator (FNO) to learn the Naiver-Stokes equations \cite{li2020fourier}. The main difference between most existing DNNs and the FNO is that the FNO is parameterized in Fourier space. Since the inputs and outputs of (partial differential equations) PDEs are continuous functions, it is more efficient to represent them in Fourier space. Benefited from the expressive and efficient architecture, the FNO achieves state-of-the-art prediction accuracy at approximating the PDEs \cite{li2020fourier}. Moreover, since convolution in the Euclidean space is equivalent to the pointwise multiplication in the Fourier space, training FNO can be significantly faster than existing DNNs. \textcolor{black}{Table \ref{tab:method} briefly summarizes the pros and cons of the popular neural network architectures.} Wen et al. extended the FNO-based architecture to simulate the multiphase flow problem \cite{wen2022u}. 
Zhijie et al. proposed an implicit U-Net enhanced Fourier neural operator (IU-FNO) for stable and efficient predictions on the long-term large-scale dynamics of 3D turbulence, reducing the prediction errors caused by temporal accumulation \cite{li2023long}. Guibas et al. proposed the Adaptive Fourier Neural Operator (AFNO) as an efficient token mixer that learns to mix in the Fourier domain. The AFNO is highly parallel with a quasi-linear complexity and has linear memory in the sequence size, and it outperforms the standard self-attention mechanisms for few-shot segmentation in terms of both efficiency and accuracy \cite{guibas2021adaptive}. Jaideep et al. applied the AFNO to predict the weather at high-resolution and fast-timescale. They proposed the FourCastNet model, which matches the forecasting accuracy of the European Center for Medium-Range Weather Forecasting (ECMWF) Integrated Forecasting System (IFS), a state-of-the-art Numerical Weather Prediction (NWP) model, at short lead times for large-scale variables \cite{pathak2022fourcastnet}.

\begin{table}[ht]
\centering
\caption{\textcolor{black}{Comparison of different network architectures}}
\label{tab:method}
\footnotesize
\begin{tabular}{lll}
\hline
&
  Pros &
  Cons \\ \hline
MLP &
  \begin{tabular}[c]{@{}l@{}} Simplicity and ease of implementation.\cite{goodfellow2016deep} \\ Good performance for approximating simple PDEs.\cite{golak2007mlp}\end{tabular} &
  \begin{tabular}[c]{@{}l@{}}Lack the mechanism to handle structured input. \cite{gardner1998artificial}\\Fail to learn complex spatial interactions. \cite{murtagh1991multilayer} \end{tabular} \\ \hline
GNN &
  \begin{tabular}[c]{@{}l@{}}Ability to model graph-structured data. \cite{zhou2020graph}\\ Effective for problems with irregular domains.\cite{li2020neural}\end{tabular} &
  \begin{tabular}[c]{@{}l@{}}Computational cost directly scales with the \\number of edges.\cite{zhou2020graph}\\ Performance may degrade for very large graphs.\cite{xu2018powerful}\end{tabular} \\ \hline
PINN &
  \begin{tabular}[c]{@{}l@{}}Physics constrained, ensuring physical consistency. \cite{raissi2019physics}\\ Less training data demand. \cite{cuomo2022scientific}\end{tabular} &
  \begin{tabular}[c]{@{}l@{}}Require prior knowledge of PDE structure. \cite{raissi2019physics}\\ Increase of training cost. \cite{raissi2019physics}\end{tabular} \\ \hline
Transformers &
  \begin{tabular}[c]{@{}l@{}}Good at capturing sequential relationship \\(e.g. temporal sequence) between elements.\cite{vaswani2017attention}
  \\ Enough model complexity to deal 
  \\with large amount of data.\cite{brown2020language} \end{tabular} &
  \begin{tabular}[c]{@{}l@{}}Computational expensive in both time and space. \cite{brown2020language}\end{tabular} \\ \hline
FNO &
  \begin{tabular}[c]{@{}l@{}}Computational efficient and quasi-linear complexity.\cite{li2020fourier}\\ Consistent error rates among different resolutions.\cite{li2020fourier} \end{tabular} &
  \begin{tabular}[c]{@{}l@{}}Error rises with higher Reynolds numbers.\cite{li2020fourier,peng2022attention} 
  \end{tabular} \\ \hline
\end{tabular}
\end{table}

In addition to FourCastNet, several machine learning methods have been recently applied to weather forecasting. 
Bi et al. introduced Pangu-Weather \cite{bi2022pangu}, a global weather forecasting system that has shown superior performance compared to traditional numerical weather prediction methods in terms of both accuracy and speed. The system utilizes a 3D earth-specific transformer architecture to incorporate height information and employs hierarchical temporal aggregation to mitigate cumulative forecast errors.
Lam et al. proposed GraphCast \cite{lam2022graphcast}, an autoregressive model based on graph neural networks and a novel high-resolution multi-scale mesh representation. In most evaluated cases, GraphCast outperformed HRES, ECMWF's deterministic operational forecasting system, and other previous ML-based weather forecasting models.
To address the limitation of generalization ability in previous data-driven methods, ClimaX \cite{nguyen2023climax} was developed. It offers a solution where the pretrained model can be fine-tuned to handle a wide range of climate and weather tasks. ClimaX can be trained using diverse datasets. It extends the Transformer architecture with innovative encoding and aggregation blocks.
Focusing on global medium-range weather forecasting, FengWu \cite{chen2023fengwu} was recently introduced from a multi-modal and multi-task perspective. It employs a deep learning architecture with model-specific encoder-decoders and a cross-modal fusion Transformer. FengWu successfully extends high accuracy global medium-range weather forecasting up to 10.75 days lead time.
Compared to the transformer and GNN models utilized in these weather forecasting approaches, FNO demands the least amount of graphics memory for model training. In the context of urban microclimate prediction, our paper demonstrates that FNO enables the training of a highly accurate 3D wind prediction model using just one Tesla A100 GPU.

\section{Methodology}

\textcolor{black}{The utilization of the FNO model for predicting urban velocity magnitude follows a straightforward workflow. To train a model for predicting the 3D wind field at fixed locations in a target city, absolute velocity magnitude 3D wind fields for continuous time intervals must first be generated. The grid resolution of the 3D data is not constrained, as the FNO model is capable of processing input data at different resolutions. Next, the training data is divided into training samples based on the desired input and output time steps, and the FNO model is trained using this data. After completing the training process, the model can be deployed to forecast the 3D wind field for the target city at fixed locations.}

\textcolor{black}{In this section, we begin by illustrating the process of data generation using a numerical solver. Subsequently, we outline the steps involved in preparing this data for machine learning model training and testing. Finally, we present the architecture of the FNO model employed in this study.}

\subsection{CityFFD simulation and data description}\label{CityFFD simulation and data description}

The training and testing data are generated from the computational fluid dynamics (CFD) simulation of the urban area by CityFFD \cite{mortezazadeh2019cityffd}. CityFFD is based on the semi-Lagrangian approach and fractional stepping method running on the graphics processing unit (GPU) to simulate urban microclimate features for modeling large-scale urban problems \cite{mortezazadeh2019slac,mortezazadeh2020solving}. The fundamental theory of CityFFD is described in the previous literature \cite{mortezazadeh2022cityffd,mortezazadeh2019adaptive,mortezazadeh2017high}. In previous studies, CityFFD is applied in urban cases to evaluate the urban wind\cite{mortezazadeh2022estimating} and thermal scenarios \cite{katal2022urban,katal2019modeling}. CityFFD can be coupled with WRF for urban wind distribution \cite{wang2023evaluating,mortezazadeh2021integrating}, and CityFFD can also be integrated with building energy simulation \cite{luo2022data}. CityFFD solves the following conservation equations in equation (\ref{eq:data}), where $\vec{U}$, $\theta$, and $p$, $Re$, $Gr$, $Pr$, $v_t$, and $\alpha_t$ are the velocity, temperature, pressure, Reynolds number, Grashof number, Prandtl number, turbulent viscosity, and turbulent thermal diffusivity, respectively. 
\begin{equation}
\label{eq:data}
\begin{aligned}
    & \nabla \cdot \vec{U} = 0 \\
    & \frac{\partial \vec{U}}{\partial t} + \left ( \vec{U} \cdot \nabla \right ) \vec{U} = - \nabla p + \left ( \frac{1}{Re} + \nu_t \right ) \nabla^2 \vec{U} - \frac{Gr}{Re^2} \theta \\
    & \frac{\partial \theta}{\partial t} + \left ( \vec{U} \cdot \nabla \right ) \theta = \left ( \frac{1}{Re \cdot Pr} +\alpha_t \right ) \nabla^2 \theta\\
\end{aligned}
\end{equation}


The advection terms of equation (\ref{eq:data}) are solved by using the Lagrangian approach: to determine the values for the unknown variables (e.g., velocity and temperature) at the position $S_c^{n+1}$, we need to compute the values of $U$ and $\theta$ at the position $S_c^n$, as shown in equation (\ref{eq:n}).
\begin{equation}
\label{eq:n}
    \mathrm{d} S_c = \vec{U} \mathrm{d} t \rightarrow S_c^n \approx S_c^{n+1} - \vec{U}\Delta t
\end{equation}
 The large eddy simulation (LES) turbulence model is then applied to capture the flow's turbulence behavior. Based on this model, turbulent viscosity is calculated by equation (\ref{eq:turbulence}).
\begin{equation}
\label{eq:turbulence}
v_t = (c_s \Delta)^2 |S|
\end{equation}
In the equation, $c_s$, $\Delta$, and $S$ are the Smagorinsky constant, filter scale, and the large-scale strain rate, respectively. The Smagorinsky constant is mostly in the range of $0.1 \le c_s \le 0.24$. CityFFD is also equipped with a 4th-order interpolation scheme to model airflow on the coarse grids and overcome the high dissipation errors. Details of the proposed method have been comprehensively investigated in previous literature \cite{mortezazadeh2019cityffd}. 

CityFFD has been well validated by several CFD benchmarks and represented acceptable accuracy for modeling an urban microclimate. The validation cases for CityFFD cover the simulation for isothermal conditions and non-isothermal conditions \cite{mortezazadeh2019cityffd,mortezazadeh2022cityffd}. With  conisdering wind speed with vertical profile, wind direction, turbulence viscosity, air temperature, ground temperature, and building surface temperature as boundary conditions, CityFFD can show good performance in both predicting wind speed and air temperature.

\begin{figure}[ht]
    \centering
    \includegraphics[width=0.5\linewidth]{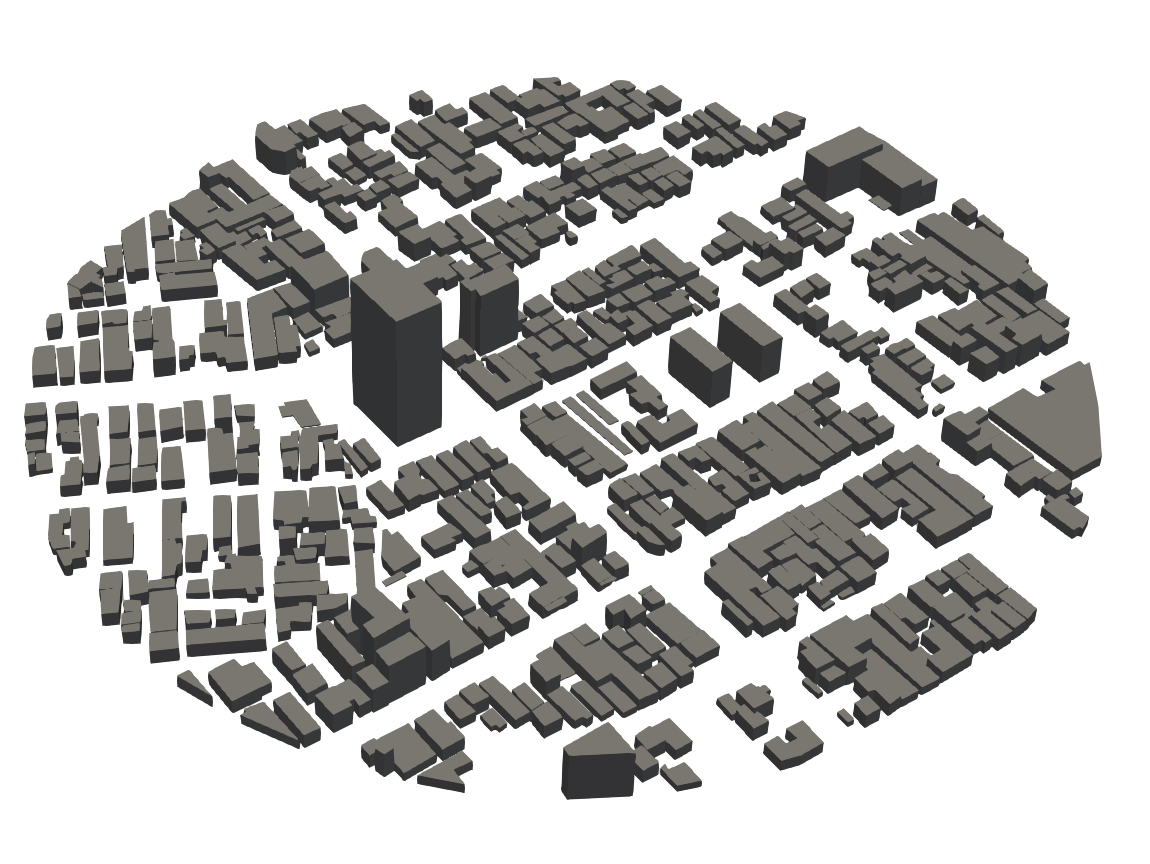}
    \caption{3D model for Region of interest in Niigata City}
    \label{fig:geometry}
\end{figure}

\begin{figure}[ht]
    \centering
    \includegraphics[width=1.0\linewidth]{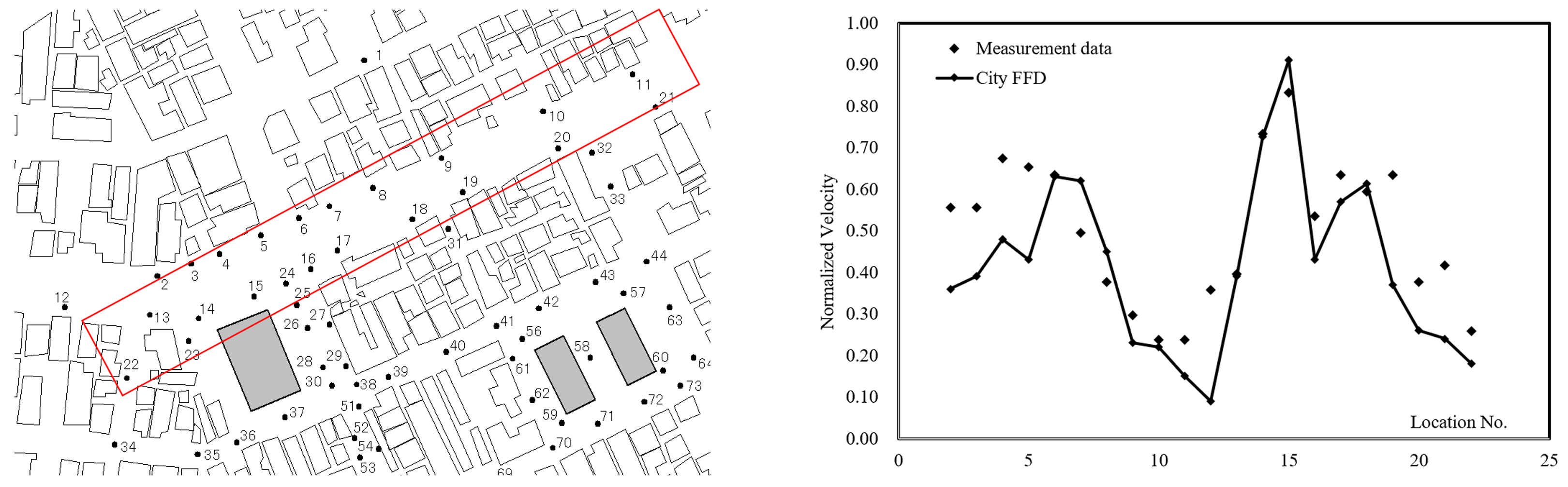}
    \caption{\textcolor{black}{CityFFD validation locations and results}}
    \label{fig:val}
\end{figure}

In this study, the performance of CityFFD is validated using a large city region in Niigata, Japan, as studied by previous researchers \cite{tominaga2008aij}. The target region, with a diameter of 400 m, is depicted in Figure \ref{fig:geometry}. The computational domain is a square with a dimension of a square of 800 m, and the elevation is at a range of 300 m. A spatial resolution of 2 m with a uniform structured grid is used, resulting in a total of 24 million grid cells. Wind speeds from the west are used for validation \cite{mortezazadeh2022cityffd}. \textcolor{black}{The validation locations are focused on the wind path inside Niigata city, and 24 locations are selected for comparing the wind speed between CityFFD simulation and
onsite measurement data in the literature\cite{tominaga2008aij} in Figure \ref{fig:val}. The detailed validation information of CityFFD can be found in our previous publication \cite{mortezazadeh2022cityffd}.} The vertical wind velocity profile is given by a power law. In this case, the wind is set as with 200 m reference velocity, and the wind profile power law exponent is 0.25 for this geometry, as stated in measurement results in the previous literature\cite{tominaga2008aij}. The simulation result is validated in the previous study for urban airflow simulation\cite{tominaga2005cross}. 

\subsection{Data pre-processing}\label{Data pre-processing}

To construct the dataset for training the FNO model, we first simulate 1200 continuous time steps of the 3D wind field of the Niigata region.  \textcolor{black}{The time interval is 0.1 second for each time step. There is one iteration for each time step. The initial wind direction is set to the West. The convergence criteria focus on the residuals of 0.001 for monitored parameters (wind speed), the courant number is 0.4 based on the reference velocity at the inlet.}
We apply a sliding window approach to segment the continuous velocity magnitudedata into samples for machine learning. Each sample consists of 6 consecutive time steps of velocity magnitude data, wherein the initial 5 time steps serve as the input and the last time step as the output.
The sliding window operates with a stride of 2 time steps, generating a total of 598 distinct data samples. \textcolor{black}{These 598 data samples contain wind fields from an initial state to a relatively stable state. We would retain and analyze the data from all stages, including the spin-up period, since we're interested in the transient development of the flow from the initial conditions to the fully developed state. Additionally, including wind field data from all stages allows us to test how well the FNO performs in a more challenging scenario rather than solely training it on stable states.} We then randomly partition this dataset into 500 samples for training and 98 samples for testing. 

Directly using 3D velocity magnitude data with a grid size of $400\times400\times150$ requires a significant amount of graphics memory on the GPU. To accommodate one training sample in the Tesla A100 GPU, downsampling of the original data is necessary. We employ 3rd-order spline interpolation to downsample the 3D velocity magnitude data into $200 \times200 \times 150$ dimensions. This interpolation technique effectively preserves the details of the original data.

Take the 1-dimension interpolation as an example, suppose we want to find the spline curve through $n+1$ points $(x_0,y_0)$, $(x_1,y_1)$, $(x_2,y_2)$, $\dots$, $(x_n,y_n)$, where $x$ is increasing, the cubic spline $S(x)$ is defined as: 
\begin{equation}
\label{eq:spline}
S(x) =
\begin{cases}
    & C_1(x), x_0 \le x \le x_1 \\
    & \cdots \\
    & C_i(x), x_{i-1} \le x \le x_i \\
    & \cdots \\
    & C_n(x), x_{n-1} \le x \le x_n
\end{cases}
\end{equation}
where each $C_i(x)$ is a cubic function: $C_i(x) = a_i + b_i x + c_i x^2 + d_i x^3 (d_i \neq 0)$.

To construct a smooth spline, we use these boundary conditions:
\begin{equation}
\label{eq:spline_boundary}
\begin{aligned}
    & C_i(x_{i-1}) = y_{i-1} \text{ and } C_i(x_i) = y_i, i = 1, 2, \dots, n\\
    & C_i ' (x_i) = C_{i+1} ' (x_i), i = 1, 2, \dots, n-1 \\
    & C_i '' (x_i) = C_{i+1} '' (x_i), i = 1, 2, \dots, n-1 \\
    & C_i '' (x_0) = C_{n} '' (x_n) = 0 .
\end{aligned}
\end{equation}

The parameters for the cubic spline interpolation can be obtained by solving this linear system.

\subsection{FNO model construction}\label{FNO model construction}

Images in the real-world have lots of edges and stripes, so CNNs can capture them well with local convolution kernels, and they are good at learning to extract high-level semantic information from natural images \cite{krizhevsky2012imagenet}. In contrast, the solutions of Partial Differential Equations (PDEs) are continuous functions, it is more efficient to represent them in Fourier space. Recently, 
Li et al. proposed to learn the Naiver-Stokes equation by parameterizing the NN model in Fourier space instead of the Euclidean space \cite{li2020fourier}. By parameterizing the neural network in Fourier space, the FNO essentially mimics the pseudo-spectral methods \cite{fan2019bcr,kashinath2020enforcing}. Numerical experiments show that the FNO achieves state-of-the-art accuracy at approximating the Navier-Stokes equations \cite{li2020fourier}.

For a finite collection of the observed input-output pairs, the Fourier neural operator learns the mapping from any functional parametric dependence to the solution, thus they learn an entire family of partial differential equations instead of a single equation \cite{li2020fourier}. Specifically, let $D \subset \mathbb{R}^{d}$ be a bounded, open set, and notate the target non-linear mapping as $G^{\dagger}: \mathcal{A}\rightarrow \mathcal{U}$, where $\mathcal{A}\left(D ; \mathbb{R}^{d_{a}}\right)$ and $\mathcal{U}\left(D ; \mathbb{R}^{d_{u}}\right)$ are separable Banach spaces of function taking values in $\mathbb{R}^{d_{a}}$ and $\mathbb{R}^{d_{u}}$ respectively \cite{beauzamy2011introduction}. The Fourier neural operators learns an approximation of $G^{\dagger}$ by constructing a mapping parameterized by $\theta \in \Theta$. $G: \mathcal{A} \times \Theta \rightarrow \mathcal{U}.$ The optimal parameters $\theta^{\dagger} \in \Theta$ are determined in the test-train setting by using a data-driven empirical approximation, such that $G\left(\cdot, \theta^{\dagger}\right)=G_{\theta^{\dagger}} \approx G^{\dagger}$. 

\begin{figure}[ht]
	\centering
	\includegraphics[width=0.7\linewidth]{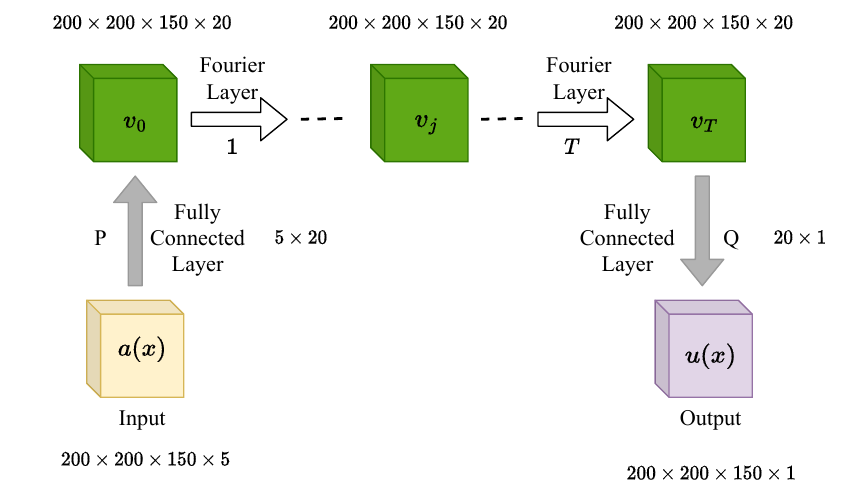}
	\caption{The Fourier neural operator architecture}
	\label{fno_architecture}
\end{figure}

The neural operators \cite{li2020neural} are formulated as an iterative architecture $v_{0} \mapsto$ $v_{1} \mapsto \ldots \mapsto v_{T}$ where $v_{j}$ for $j=0,1, \ldots, T-1$ is a sequence of functions each taking values in $\mathbb{R}^{d_{v}}$. The overall architecture of Fourier Neural Operator (FNO) is shown in figure \ref{fno_architecture}.

First, the input $a(x)$ is lifted to a higher dimension channel space by a fully connected layer $P$, where $P$ is a fully connected layer parameterized by a $5\times20$ real-valued weight matrix. Then a sequence of Fourier layers is applied, where the Fourier layer keeps the tensor shape unchanged. Finally the output $u(x)$ is obtained by projecting back to the target dimension by another fully connected layer $Q$, where $Q$ is a fully connected layer parameterized by a $20\times1$ real-valued weight matrix.

The higher dimensional representation is updated iteratively through a
sequence of Fourier layers, as shown in equation (\ref{eq:update}), where $\mathcal{K}:\mathcal{A} \times \Theta_{\mathcal{K}} \rightarrow \mathcal{L}\left(\mathcal{U}\left(D ; \mathbb{R}^{d_{v}}\right), \mathcal{U}\left(D ; \mathbb{R}^{d_{v}}\right)\right)$ maps to bounded linear operators on $\mathcal{U}\left(D ; \mathbb{R}^{d_{v}}\right)$ and is parameterized by $\phi \in \Theta_{\mathcal{K}}$, $W: \mathbb{R}^{d_{v}} \rightarrow \mathbb{R}^{d_{v}}$ is a linear transformation, and $\sigma: \mathbb{R} \rightarrow \mathbb{R}$ is an elementally defined non-linear activation function.
\begin{equation}\label{eq:update}
v_{t+1}(x) =\sigma\left(W v_{t}(x)+\left(\mathcal{K}(a ; \phi) v_{t}\right)(x)\right), \quad \forall x \in D
\end{equation}

Let $\mathcal{F}$ and $\mathcal{F}^{-1}$ denote the Fourier transform and its inverse transform of a function $f: D \rightarrow \mathbb{R}^{d_{v}}$ respectively. Replacing the kernel integral operator in equation (\ref{eq:update}) by a convolution operator defined in Fourier space, and applying the convolution theorem, the Fourier integral operator can be expressed by equation (\ref{eq:k}), where $R_{\phi}$ is the Fourier transform of a periodic function $\kappa: \bar{D} \rightarrow \mathbb{R}^{d_{v} \times d_{v}}$ parameterized by $\phi \in \Theta_{\mathcal{K}}$. 
\begin{equation}\label{eq:k}
\left(\mathcal{K}(\phi) v_{t}\right)(x)=\mathcal{F}^{-1}\left(R_{\phi} \cdot\left(\mathcal{F} v_{t}\right)\right)(x), \quad \forall x \in D
\end{equation}

The frequency mode $k \in D$ is assumed to be
periodic, and it allows a Fourier series expansion, which expresses as the discrete modes $k \in \mathbb{Z}^{d}$. The finite-dimensional parameterization is implemented by truncating the Fourier series at a maximal number of modes $k_{\max }=\left|Z_{k_{\max }}\right|=\mid\left\{k \in \mathbb{Z}^{d}:\left|k_{j}\right| \leq k_{\max , j}\right.$, for $\left.j=1, \ldots, d\right\} \mid$. We discretize the domain $D$ with  $n \in \mathbb{N}$ points, where $v_{t} \in \mathbb{R}^{n \times d_{v}}$ and $\mathcal{F}\left(v_{t}\right) \in \mathbb{C}^{n \times d_{v}}$. $R_{\phi}$ is parameterized as complex-valued weight tensor containing a collection of truncated Fourier modes $R_{\phi} \in \mathbb{C}^{k_{\max } \times d_{v} \times d_{v}}$, and $\mathcal{F}\left(v_{t}\right) \in \mathbb{C}^{k_{\max } \times d_{v}}$ is obtained by truncating the higher modes, therefore $\left(R_{\phi}\cdot\left(\mathcal{F} v_{t}\right)\right)_{k, l}=\sum_{j=1}^{d_{v}} R_{\phi  k, l, j}\left(\mathcal{F} v_{t}\right)_{k, j}, \quad k=1, \ldots, k_{\max }, \quad j=1, \ldots, d_{v}.$

In CFD modeling, the flow is typically uniformly discretized with resolution $s_{1} \times \cdots \times s_{d}=n$, and $\mathcal{F}$ can be replaced by the Fast Fourier Transform (FFT). For $f \in \mathbb{R}^{n \times d_{v}}, k=\left(k_{1}, \ldots, k_{d}\right) \in \mathbb{Z}_{s_{1}} \times \cdots \times \mathbb{Z}_{s_{d}}$, and $x=\left(x_{1}, \ldots, x_{d}\right) \in D$ the FFT $\hat{\mathcal{F}}$ and its inverse $\hat{\mathcal{F}}^{-1}$ are given by equation. (\ref{eq:fft}), for $l=1, \ldots, d_{v}$.
\begin{equation}\label{eq:fft}
\begin{aligned}
&(\hat{\mathcal{F}} f)_{l}(k)=\sum_{x_{1}=0}^{s_{1}-1} \cdots \sum_{x_{d}=0}^{s_{d}-1} f_{l}\left(x_{1}, \ldots, x_{d}\right) e^{-2 i \pi \sum_{j=1}^{d} \frac{x_{j} k_{j}}{s_{j}}}. \\
&\left(\hat{\mathcal{F}}^{-1} f\right)_{l}(x)=\sum_{k_{1}=0}^{s_{1}-1} \cdots \sum_{k_{d}=0}^{s_{d}-1} f_{l}\left(k_{1}, \ldots, k_{d}\right) e^{2 i \pi \sum_{j=1}^{d} \frac{x_{j} k_{j}}{s_{j}}}
\end{aligned}
\end{equation}

In this 3D dynamic urban microclimate simulation task, the FNO takes input tensor $a(x)$ of shape $200\times200\times150\times5$, where the first three dimension $200\times200\times150$ is the grid mesh in the spatial $x,y,z$ direction, and the last dimension $5$ is the number of sequential time steps. The input $a(x)$ is lifted to a higher dimension channel space by a fully connected layer $P$, where $P$ is a fully connected layer parameterized by a $5\times20$ real-valued weight matrix.

\begin{figure}[ht]
	\centering
	\includegraphics[width=\linewidth]{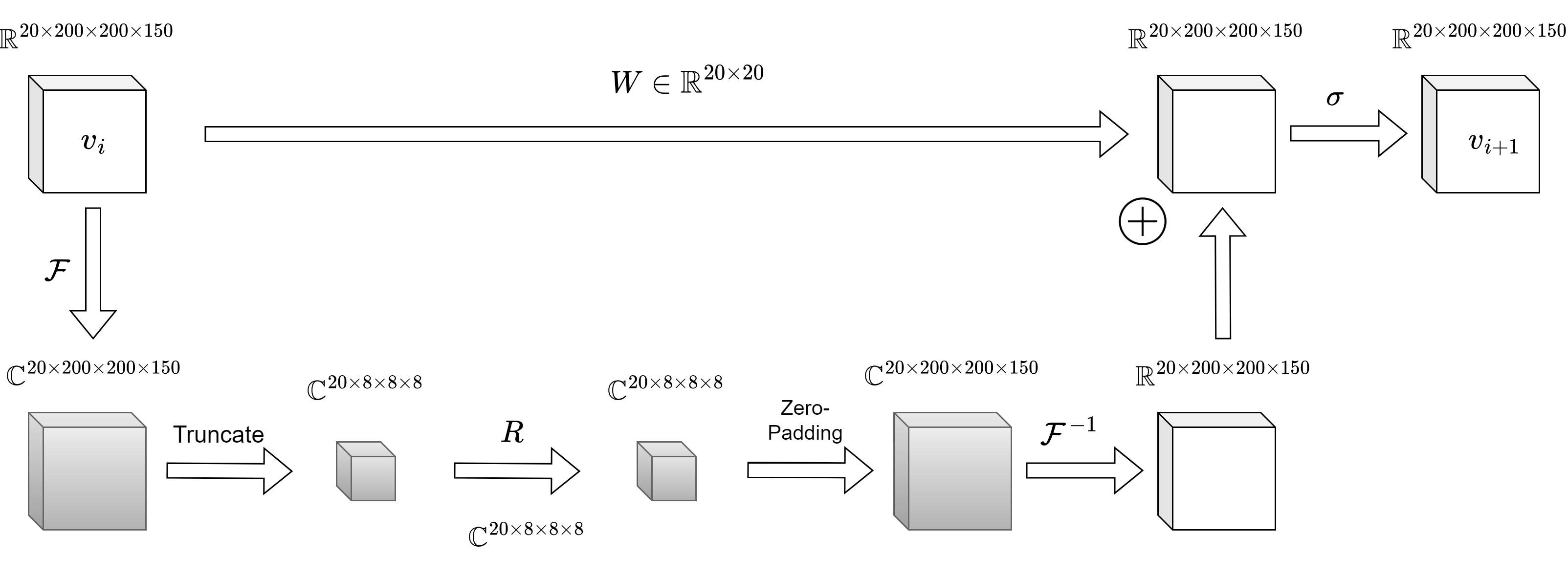}
	\caption{The Fourier layer.}
	\label{fourier_layer}
\end{figure}

The core of FNO is the sequence of Fourier layers \cite{li2020fourier}, as shown in figure \ref{fourier_layer}. The Fourier layer adopts an encoder-decoder structure, and it mainly consists of three operations: 1. Fourier transform $\mathcal{F}$; 2. linear transform on the truncated Fourier modes $R$; 3. inverse Fourier transform $\mathcal{F}^{-1}$. 

The first Fourier layer takes $v_{i}\in \mathbb{R}^{200\times200\times150\times20}$ as input tensor, and $v_{i}$ is permuted into $20\times200\times200\times150$ before the Fourier transformation. Then the higher modes in the frequency space are truncated to obtain $\mathcal{F}(v_{i})\in \mathbb{C}^{20\times8\times8\times8}$. The weight matrix $R$ is parameterized by a $20\times8\times8\times8$ complex-valued tensor. $\mathcal{F}(v_{i})R$ is zero-padded into $20\times200\times200\times150$ complex-valued matrix before the inverse Fourier transform $\mathcal{F}^{-1}$. The non-linear activation is applied after the inverse Fourier transform.

\section{Results and discussion}
In this section, we assess the reconstruction performance of the FNO model over 3D dynamic urban microclimate data. We begin by evaluating the one-step prediction accuracy of the FNO model and its generalization ability across different criteria. These criteria include the distribution of velocity magnitude in the 3D wind field, the conditional average error of the velocity, and the velocity change with respect to varying heights. Subsequently, we examine the accumulated error of the FNO model by testing its predictive accuracy over sequential time steps. To facilitate the comparison between the ground truth and the FNO model's predictions of the 3D wind field data, we present the results using both 2D slices and 3D isosurfaces.

\subsection{Model training}

Due to instances where the absolute velocity magnitude is 0 at certain grids, the typical grid-wise relative error can't be directly computed. Therefore, we adopt the average layer-wise relative error as the loss function. We first split the 3D velocity magnitude field into 2D slices. For each slice, we compute the 2-norm of the velocity. Let $v_{x,y,z}$ and $u_{x,y,z}$ represent the ground truth and predicted velocities at grid $(x,y,z)$, respectively, and let $n_z$ denote the number of $z$ grids.
The velocity magnitude norm for layer $z$, denoted as ${|v_z|}_2$, is computed as $\sqrt{\sum_{x,y}v_{x,y,z}^2}$. We then compute the relative error of this layer by dividing the 2 norm of the error with the corresponding velocity magnitude norm. Finally, we calculate the average relative error per layer to obtain the overall error:

\begin{equation}
\label{eq:loss}
Loss = 
\frac{1}{n_z}\sum_{z} \frac{\sqrt{\sum_{x,y}\left(u_{x,y,z} - v_{x,y,z}\right)^2}}{{|v_z|}_2}
=\frac{1}{n_z}\sum_{z} \sqrt{\frac{\sum_{x,y}\left(u_{x,y,z} - v_{x,y,z}\right)^2}{\sum_{x,y}v_{x,y,z}^2}}.
\end{equation}

We utilize this average layer-wise relative error, equation (\ref{eq:loss}), as the loss function to train the FNO model.

\begin{figure}[ht]
\centering
\includegraphics[width = 0.5\textwidth]{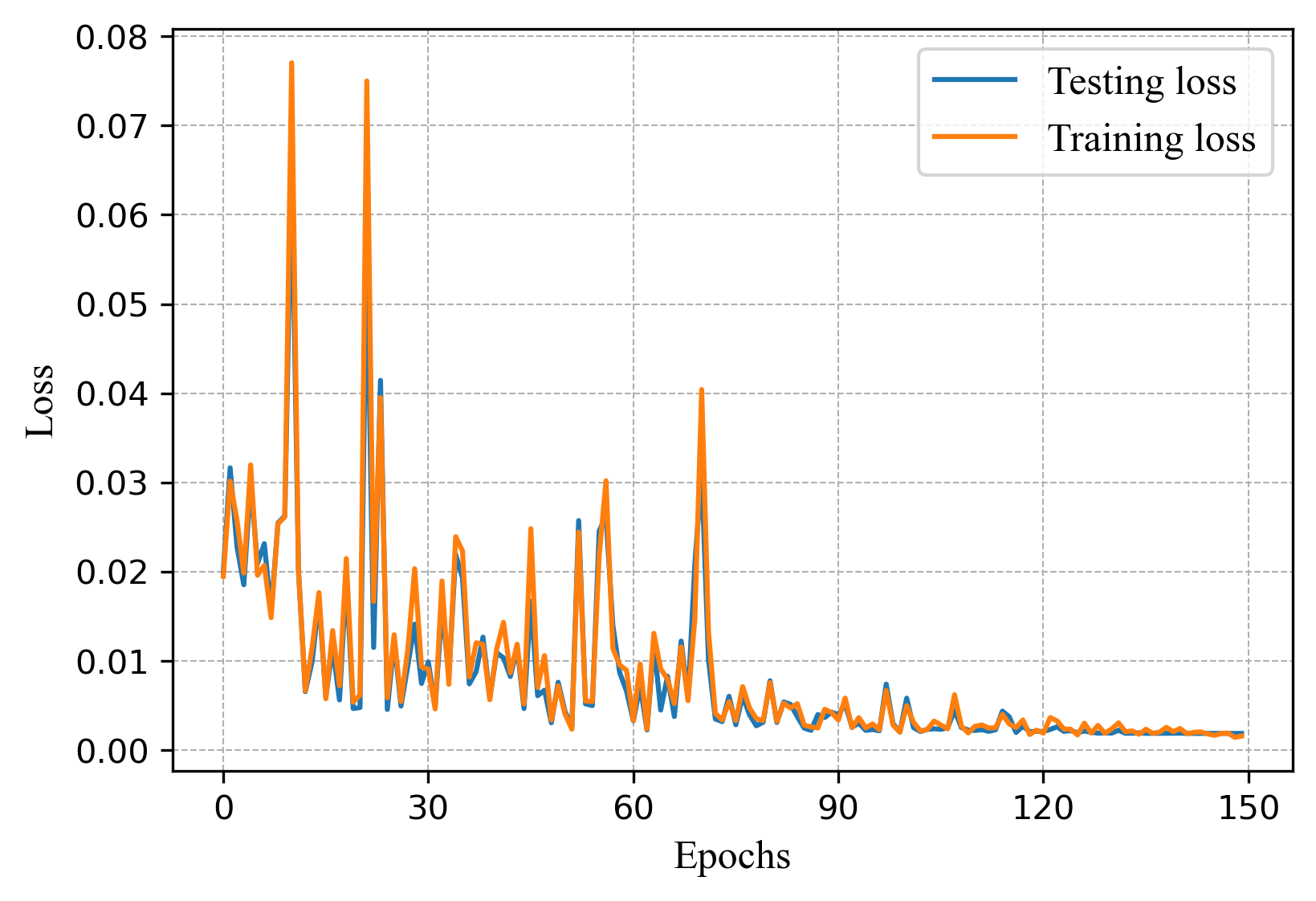}
\caption{Training and testing curve for the FNO model}
\label{fig:train_curve}
\end{figure}

In our training process, we leverage the FNO model that comprises four Fourier layers. We set the mode parameter as 8 and the width parameter as 20, as suggested in reference \cite{li2020fourier}. The FNO model is trained on the generated 3D velocity magnitude data for a total of 150 epochs, with a learning rate of 0.001 and a batch size of 1. Each epoch takes approximately 20 minutes to train on a Tesla A100 GPU. The entire training process for the FNO model spans around 50 hours. Figure \ref{fig:train_curve} illustrates the progression of the training loss and testing loss throughout the training duration.

\subsection{Testing performance : one-step prediction and generalization ability}

We first evaluate the one-step prediction error of our model. Our model is solely trained on the west wind (0°) case. The input is 5 continuous time steps of 3D wind data and the output is the next one time step of 3D wind data. \textcolor{black}{The reason why the input tensor takes 5 sequential time steps of the solution is to trade-off between memory consumption and prediction accuracy \cite{peng2023linear}.} The test loss on the west wind case demonstrates remarkable accuracy, with a value of 0.3\%.

To evaluate the model's generalization capability, we further examine its prediction accuracy in three distinct wind directions. We generate 50 continuous time steps of 3D wind field data for north wind (90°), east wind (180°), and south wind (270°), respectively. When the model is tested on winds differing by 90 degrees (north and south wind), the test loss increases to approximately 3\%, roughly tenfold compared to the West wind case. For the opposite wind direction (east wind), the test loss increases to around 5\%. These results indicate that the FNO model, trained solely on data from one wind direction, is still capable of predicting wind fields for other wind directions with an acceptable prediction error. \textcolor{black}{The FNO model exhibits an advantageous generalization ability to learn the Navier–Stokes equations because it can capture the dynamic characteristics of wind flow within the frequency domain. Moreover, the Fourier operators employed in FNO possess a global scope, covering the entire input data. In contrast, conventional machine learning models such as CNNs rely on convolutional filters to capture local spatial features, often leading to diminished generalization capability.
Additionally, in order to enhance the FNO model’s ability to capture the physical relation between its inputs and outputs, we choose a small time interval of 0.1 seconds when generating continuous time steps of wind field data. This choice allows consecutive data to preserve critical physical relationship information better. Therefore, the time interval configuration also contributes to the promising performance of the FNO model.} However, it is observed that the prediction error tends to be larger when dealing with different wind directions. The model's performance can be significantly improved by including data with various wind directions into the training dataset.

\begin{figure}[ht]
\centering
\includegraphics[width=0.9\linewidth]{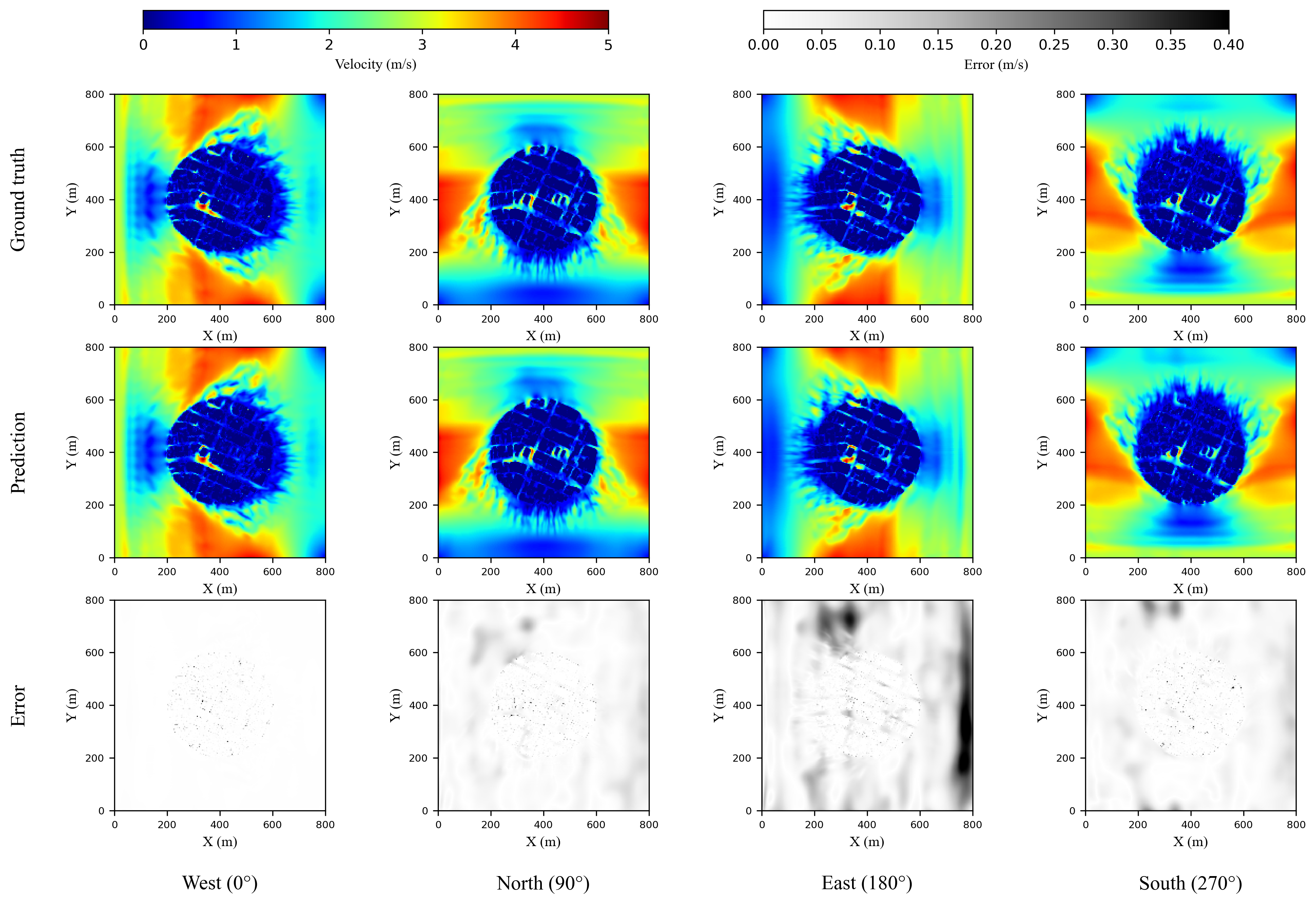}
\caption{One-step prediction comparison for different wind directions at 2m high horizontal slice \textcolor{black}{with training data of west wind only (if not explicitly stated, the training data of other figures similarly only includes west wind)}}
\label{fig:1_step_2d}
\end{figure}

Figure \ref{fig:1_step_2d} shows the 2D visualization of the ground truth, prediction, and the absolute error of one data sample for the one-step prediction, represented through a horizontal slice at a height of 2 meters. The wind field at this height is particularly relevant to people's experiences in the environment. 
In the training scenario (west wind), the absolute error depicted in the figure is minimal. \textcolor{black}{Upon conducting detailed observation, it is evident that there are no significant disparities in the errors of the predicted wind field within the streets, squares, and open spaces around the city. These errors consistently manifest at a minute scale. The error is more related to the changing speed of the wind field. The degree of error is closely associated with variations in the wind velocity. Specifically, areas characterized by transitions from low to high or high to low wind velocities tend to exhibit slightly larger errors. The prediction error peaks are predominantly located at the edges of buildings. This phenomenon arises from the inherent challenge faced by the FNO model in effectively capturing boundary information solely from its training data.} 
For winds differing by 90 degrees (north and south wind), larger errors are observed in the open spaces surrounding the buildings. For the opposite wind direction (east wind), significant errors are evident in specific areas of the open space.

\begin{figure}[htp]
\centering
\includegraphics[width=0.9\linewidth]{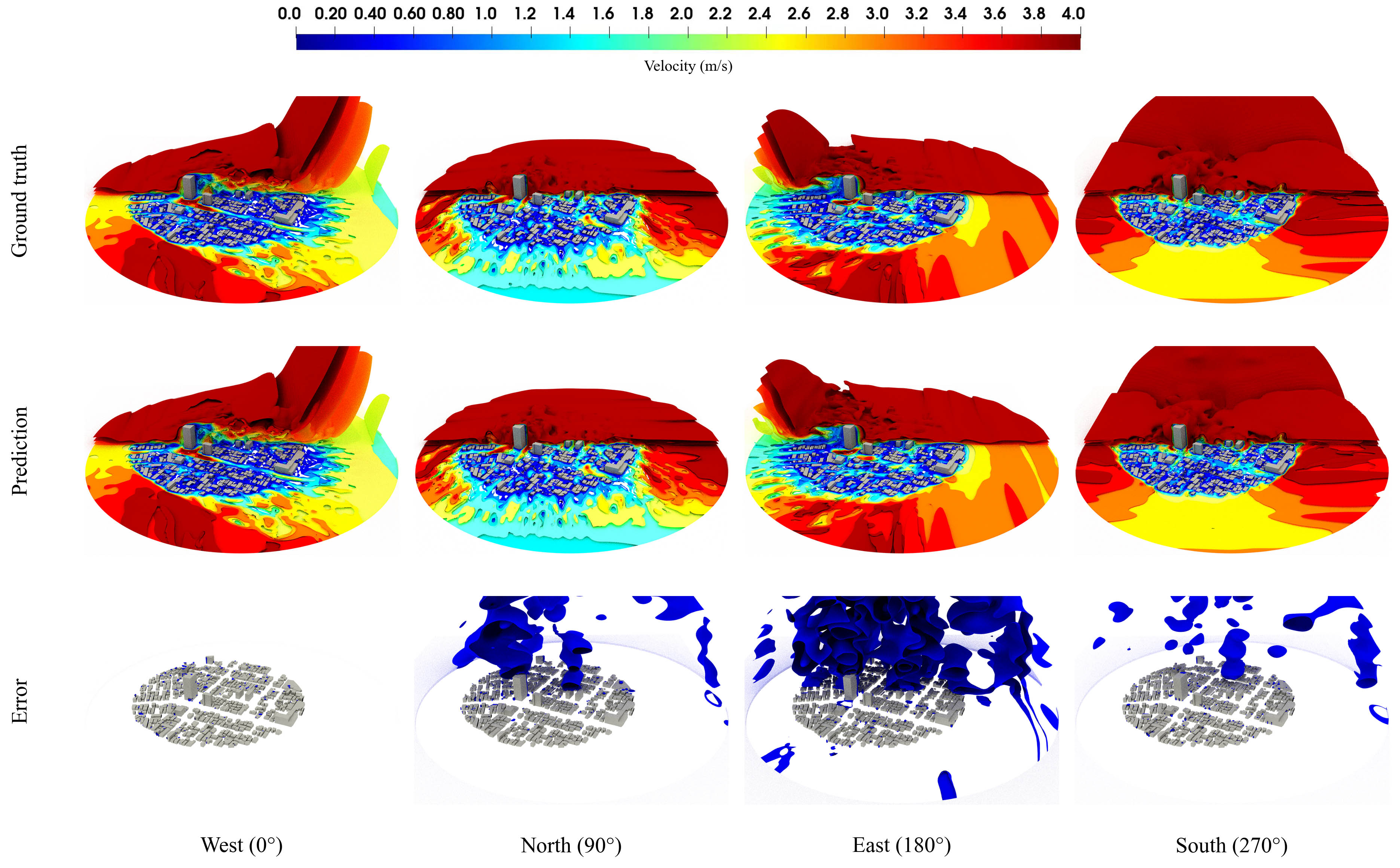}
\caption{One-step prediction comparison for different wind directions in 3D isosurfaces}
\label{fig:1_step_3d}
\end{figure}

Figure \ref{fig:1_step_3d} shows the 3D visualization of the ground truth, prediction, and the absolute error of one data sample for the one-step prediction through 3D isosurfaces. This figure presents the isosurfaces of the wind field below 4 m/s within a hemisphere region situated above the ground. To enhance the clarity of the wind field within urban areas, only the isosurfaces below 2 meters are displayed on the front half of the hemisphere. Conversely, the complete set of isosurfaces is shown on the back half of the hemisphere. The results clearly demonstrate that the training scenario (west wind) has the best prediction performance and the opposite direction (east wind) shows the largest absolute error.

\begin{figure}[htp]
\centering
\includegraphics[width=\linewidth]{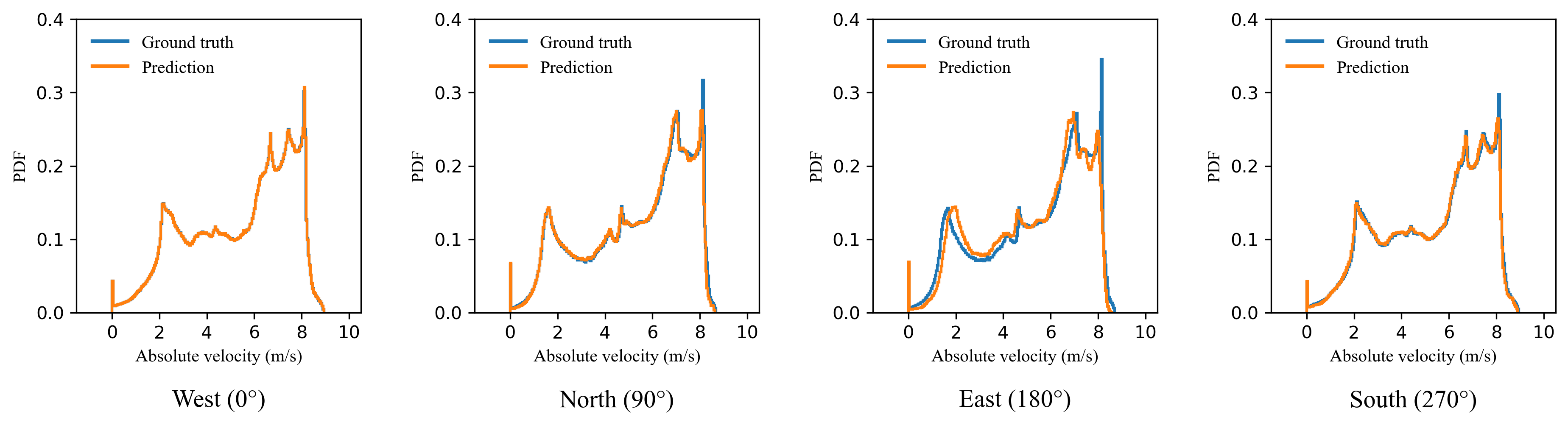}
\caption{Probability density function of the absolute velocity magnitude for different wind directions}
\label{fig:pdf}
\end{figure}

Figure \ref{fig:pdf} presents the probability density functions (PDFs) of the absolute velocity magnitude for various wind directions. 
The velocity magnitude of the wind inside the buildings is defined as 0 and is included in the PDFs as well.
In the case of the training scenario (west wind), the velocity magnitude distribution of the predicted wind field exhibits a close match with that of the ground truth. 
Similarly, for winds differing by 90 degrees (north and south wind), the velocity magnitude distribution of the prediction remains in good agreement with the ground truth. 
The most significant gap between the prediction and the ground truth is observed for the opposite wind direction (east wind). 
This gap is evident across almost the entire velocity magnitude range, from 0 to 9 m/s. 

\begin{figure}[htp]
\centering
\includegraphics[width=\linewidth]{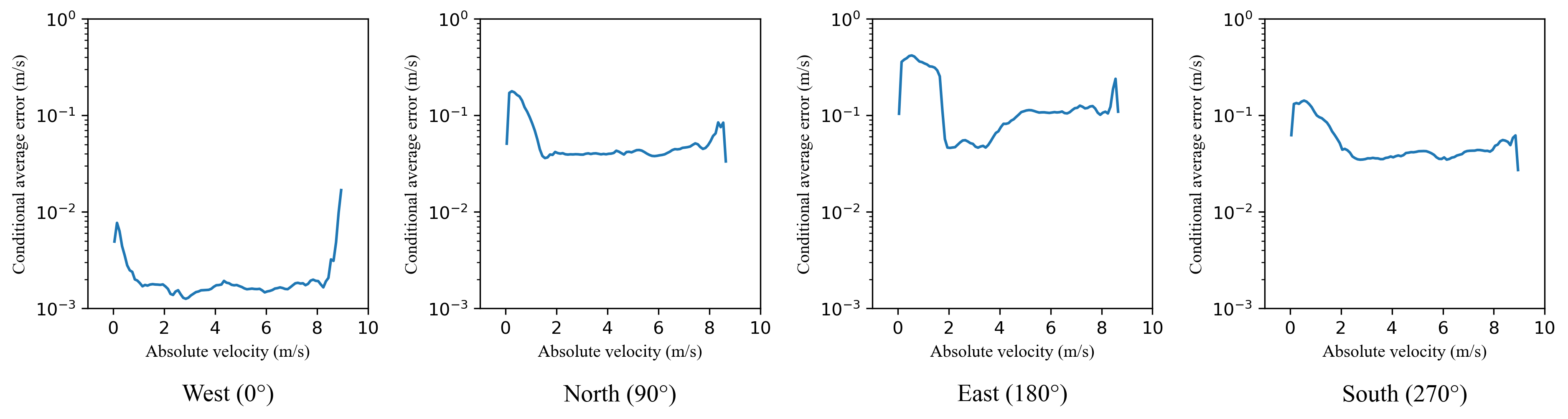}
\caption{Conditional average absolute error of the absolute velocity magnitude for different wind directions}
\label{fig:c_error}
\end{figure}

Figure \ref{fig:c_error} illustrates the conditional average absolute error at different wind directions on a logarithmic scale. As discussed earlier, the training scenario (west wind) exhibits the smallest absolute error, while the opposite direction (east wind) displays the largest absolute error. Across all wind directions, the absolute error tends to be higher for both low and high velocities. This is because the FNO is a data-driven model which lacks strict boundary constraints. Consequently, the model may encounter challenges in accurately predicting wind velocities at extreme ends of the velocity magnitude range.

\begin{figure}[htp]
\centering
\includegraphics[width=\linewidth]{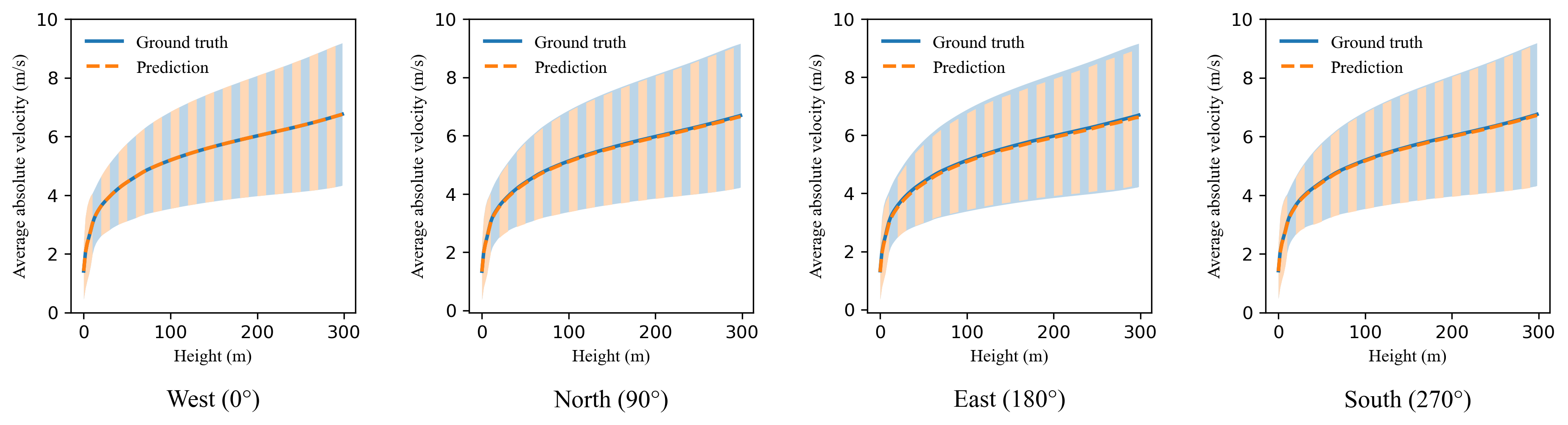}
\caption{Average absolute velocity magnitude with respect to varying heights for different wind directions}
\label{fig:height}
\end{figure}

Figure \ref{fig:height} shows the average absolute velocity magnitude as a function of varying heights for different wind directions. The corresponding colored areas represent the standard deviation of the absolute velocity. Across all wind directions, the predicted velocities closely align with the ground truth velocities at various heights.

\begin{figure}[htp]
\centering
\includegraphics[width=0.9\linewidth]{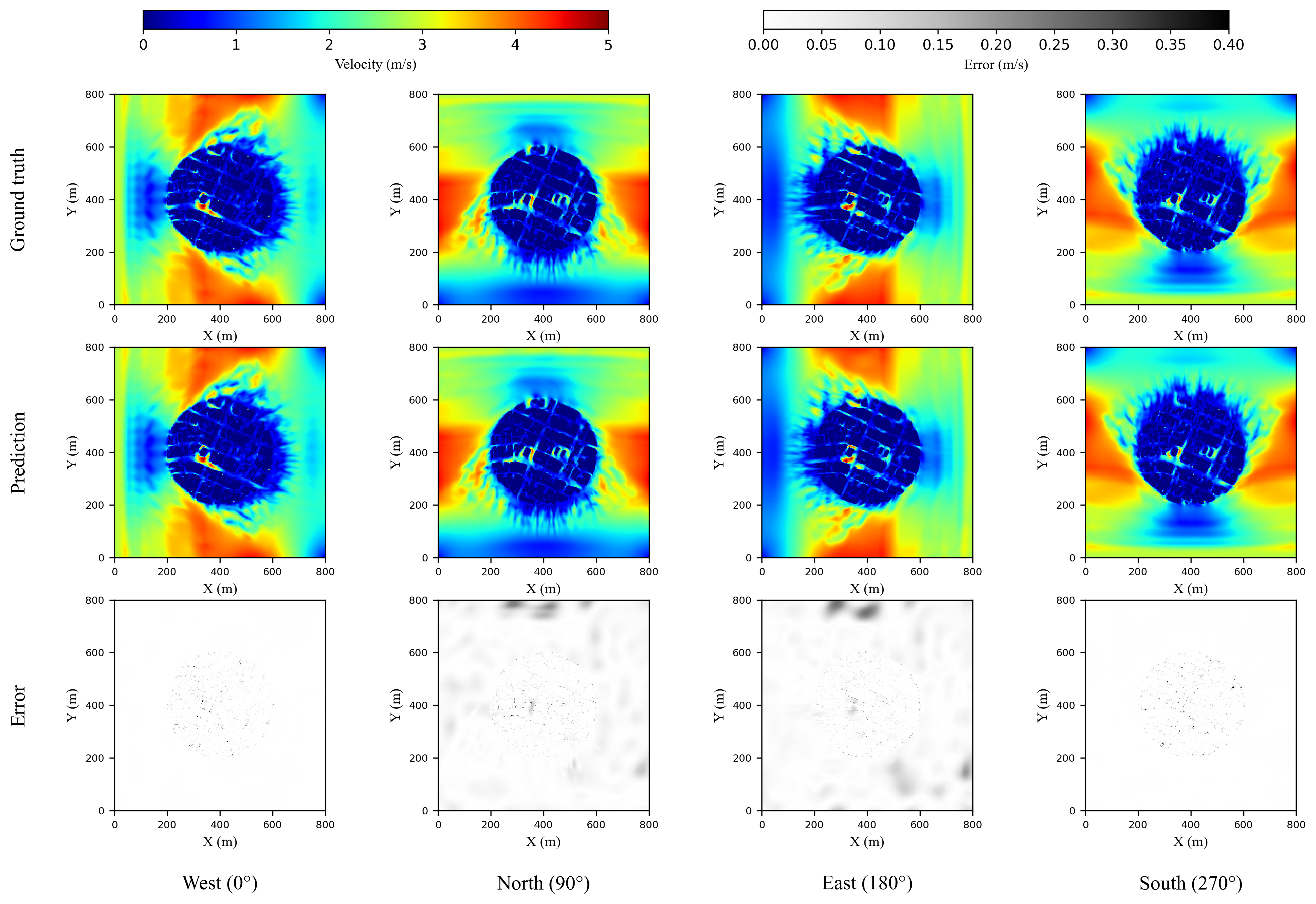}
\caption{\textcolor{black}{One-step prediction comparison for different wind directions at 2m high horizontal slice with training data of west and south wind}}
\label{fig:1_step_2d_2wind}
\end{figure}

\textcolor{black}{To further evaluate the performance of the FNO model when the training data contains more wind directions, we train the FNO model with both west wind (0°) and south wind (270°). With two directions of wind as training data, Figure \ref{fig:1_step_2d_2wind} shows the 2D visualization of the ground truth, prediction, and the absolute error of one data sample for the one-step prediction, represented through a horizontal slice at a height of 2 meters. }

\textcolor{black}{Comparing these results to those presented in Figure \ref{fig:1_step_2d}, we observed notable improvements. Specifically, the north wind (90°) test loss decreases from 3\% to 2\%, and the east wind (180°) test loss decreases from 5\% to 2\%. The inclusion of south wind (270°) in the training dataset results in a substantial reduction in its test loss, dropping from 3\% to 0.3\%. This test loss is now similar to the loss observed for west wind (0°). These results demonstrate the effectiveness of including training data of more wind directions in improving the generalization performance of the FNO model. However, the west wind (0°) test loss doesn't exhibit a further decrease, suggesting that, with sufficient training data in one wind direction, the performance of FNO in this direction reaches a limit and cannot be significantly improved by incorporating data from other directions.}

\subsection{Testing performance : sequential time step prediction}

In the context of wind field prediction with data-driven methods, errors are inherently introduced at each step of prediction. When using these prediction results to further forecast subsequent wind fields, errors tend to accumulate. Given the chaotic nature of wind fields, this accumulated error has the potential to grow exponentially. Reducing the accumulated error for sequential time step prediction remains a challenge for all data-driven wind field prediction methods.

\begin{figure}[ht]
\centering
\includegraphics[width=0.9\linewidth]{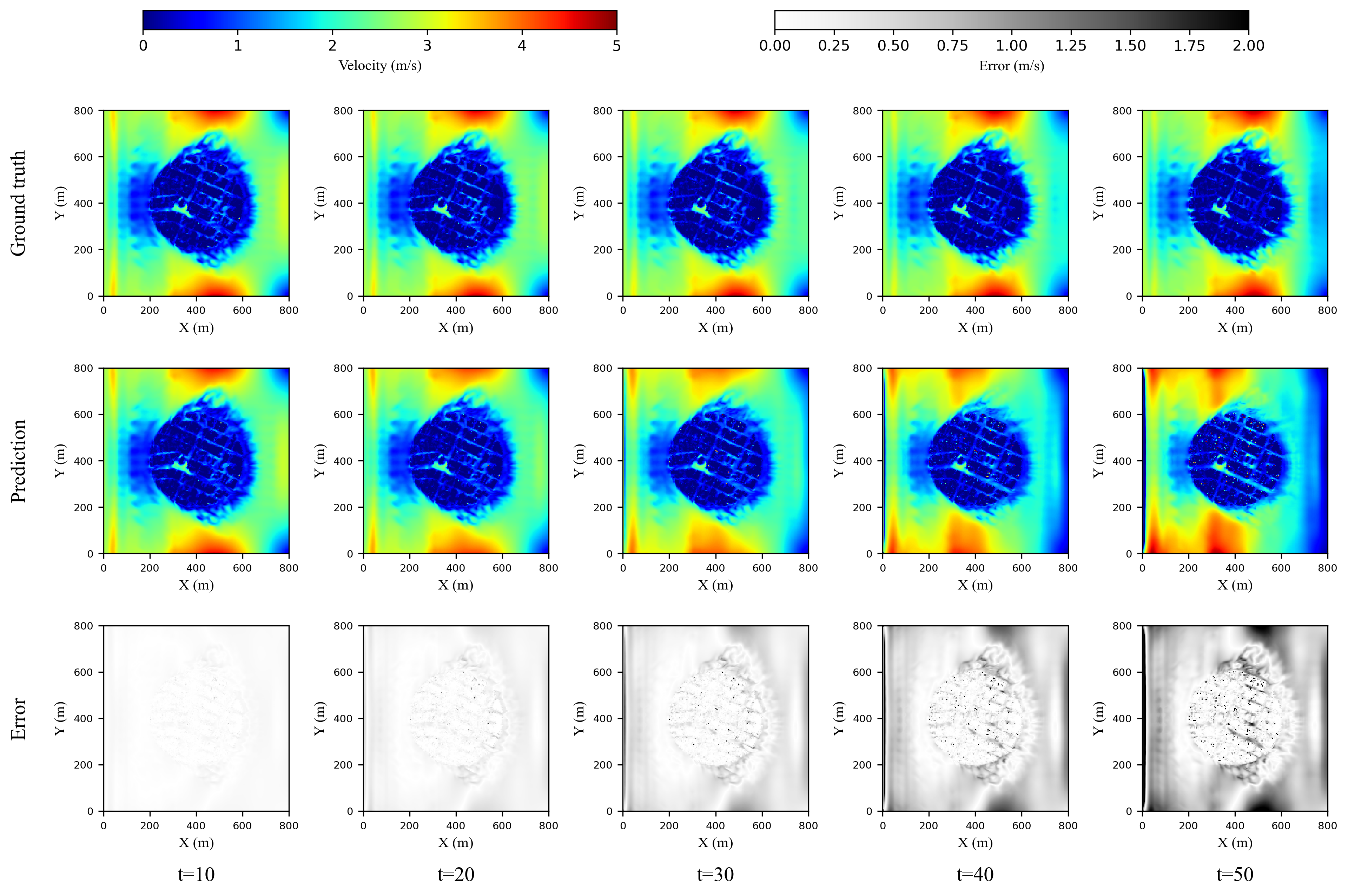}
\caption{Sequential time step prediction comparison for different wind directions at 2m high horizontal slice}
\label{fig:50_step_2h}
\end{figure}

\begin{figure}[ht]
\centering
\includegraphics[width=0.9\linewidth]{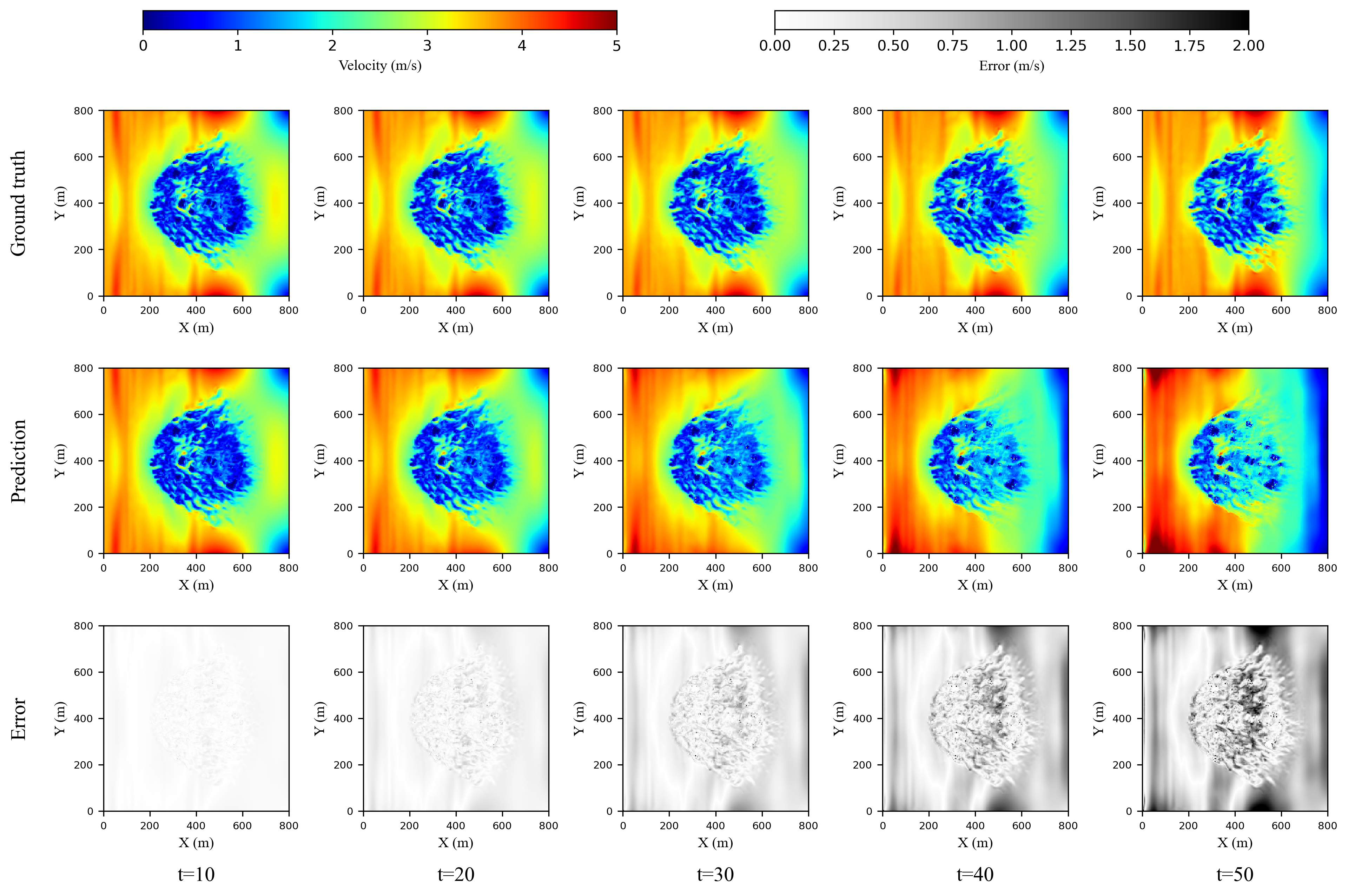}
\caption{Sequential time step prediction comparison for different wind directions at 10m high horizontal slice}
\label{fig:50_step_10h}
\end{figure}

To evaluate the accumulated error for the FNO model, we use the FNO to predict wind fields sequentially on the training scenario (west wind). With 5 initial inputs, the FNO model predicts the wind field's velocity magnitude for the 6th time step. Subsequently, this prediction is utilized as input to forecast the wind field's velocity magnitude at the 7th time step. This iterative prediction process is repeated for 50 steps, generating a sequence of predicted wind fields. These 50 predicted wind fields are then compared with the ground truth. 

Figures \ref{fig:50_step_2h} and \ref{fig:50_step_10h} show the visualization of the ground truth, prediction, and the absolute error of the accumulated error in 2D horizontal slices at heights of 2m and 10m, respectively. 
It is noted that the FNO model can accurately reconstruct the instantaneous 2D horizontal velocity magnitude field at heights of both 2m and 10m in the beginning. However, the temporal accumulated error grows larger in terms of both magnitude and region as time progresses. 
Moreover, the errors become more significant at regions where the velocity magnitude has dramatically changed, as illustrated in both figure\ref{fig:50_step_2h} and figure \ref{fig:50_step_10h}. This phenomenon is the inherent drawback of the FNO approach, which has also been reported in the recent work of applying FNO for turbulence simulations \cite{peng2022attention}.

\begin{figure}[htp]
\centering
\includegraphics[width=0.9\linewidth]{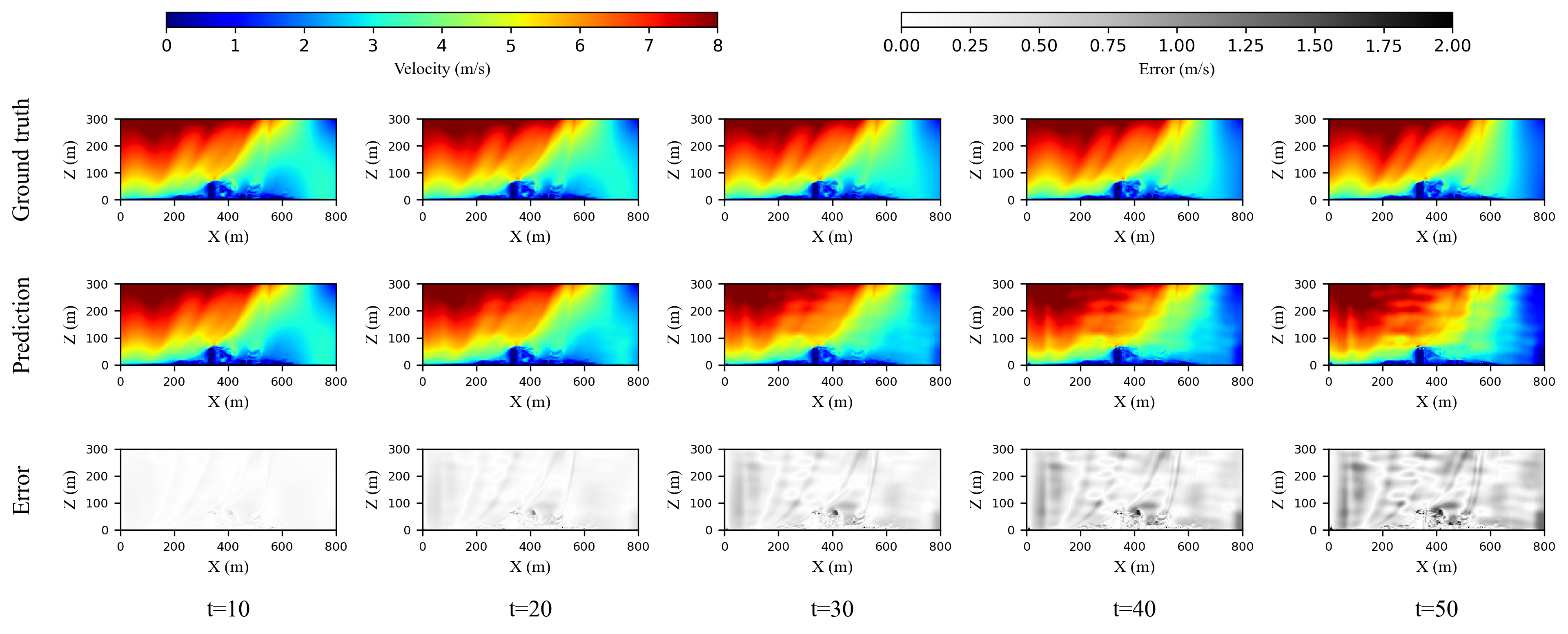}
\caption{Sequential time step prediction comparison for different wind directions at vertical slice}
\label{fig:50_step_x}
\end{figure}

Figure \ref{fig:50_step_x} visualizes the accumulated error through a 2D vertical slice. We noted that the FNO model can accurately predict the 2D vertical velocity magnitude field in the beginning, and the temporal accumulated error is enlarged 
as time progresses.

\begin{figure}[htp]
\centering
\includegraphics[width=0.9\linewidth]{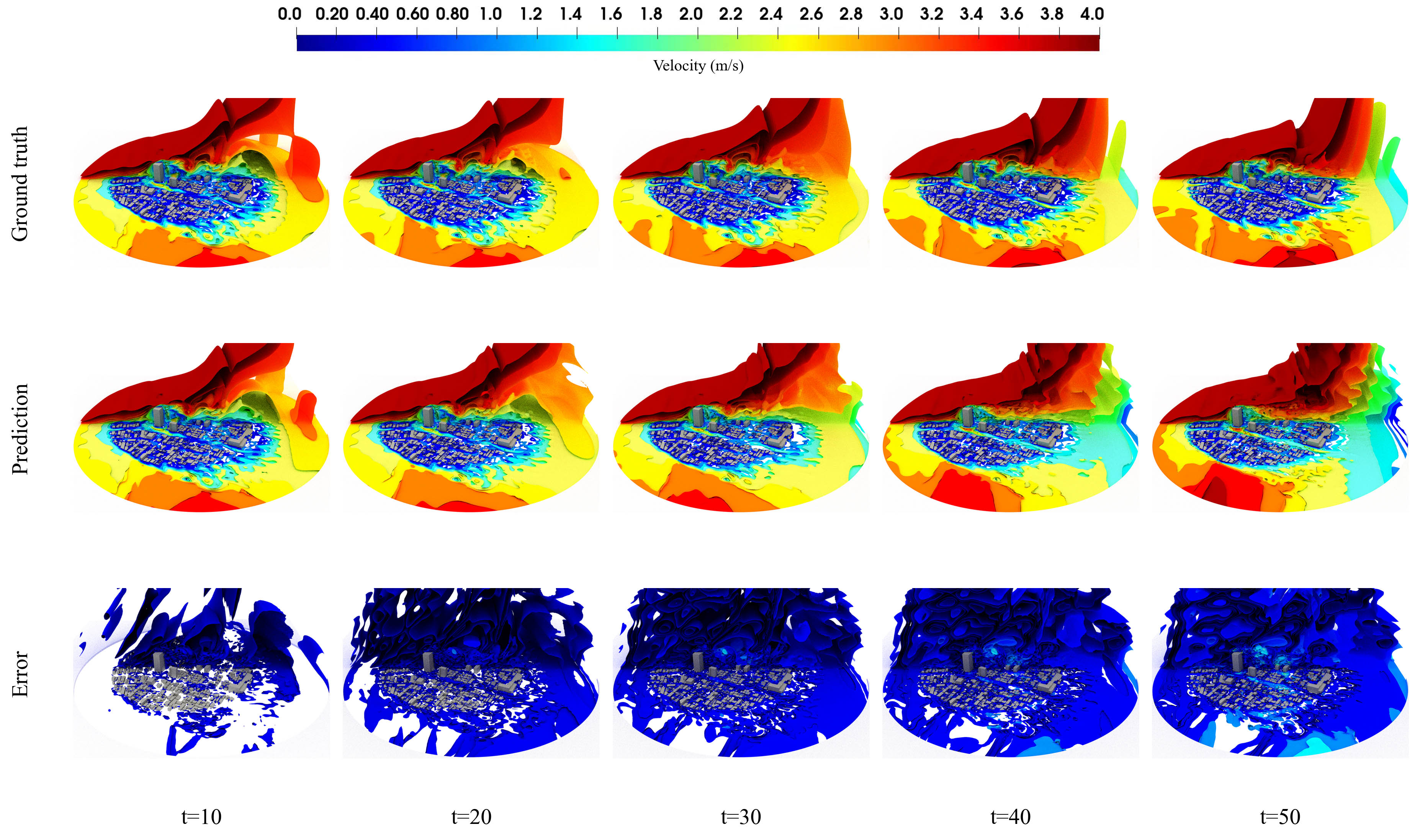}
\caption{Sequential time step prediction comparison for different wind directions in 3D isosurfaces}
\label{fig:50_step_3d}
\end{figure}

Figure \ref{fig:50_step_3d} displays the accumulated error via 3D isosurfaces within a hemisphere region above the ground. The FNO model can accurately reconstruct the instantaneous spatial 3D velocity magnitude field in the beginning. However, the temporal accumulated error grows larger in terms of both magnitude and region as time progresses. Correspondingly, in the error section of the figure, we observe a growing number of isosurfaces.

\begin{figure}[htp]
\centering
\includegraphics[width=0.5\linewidth]{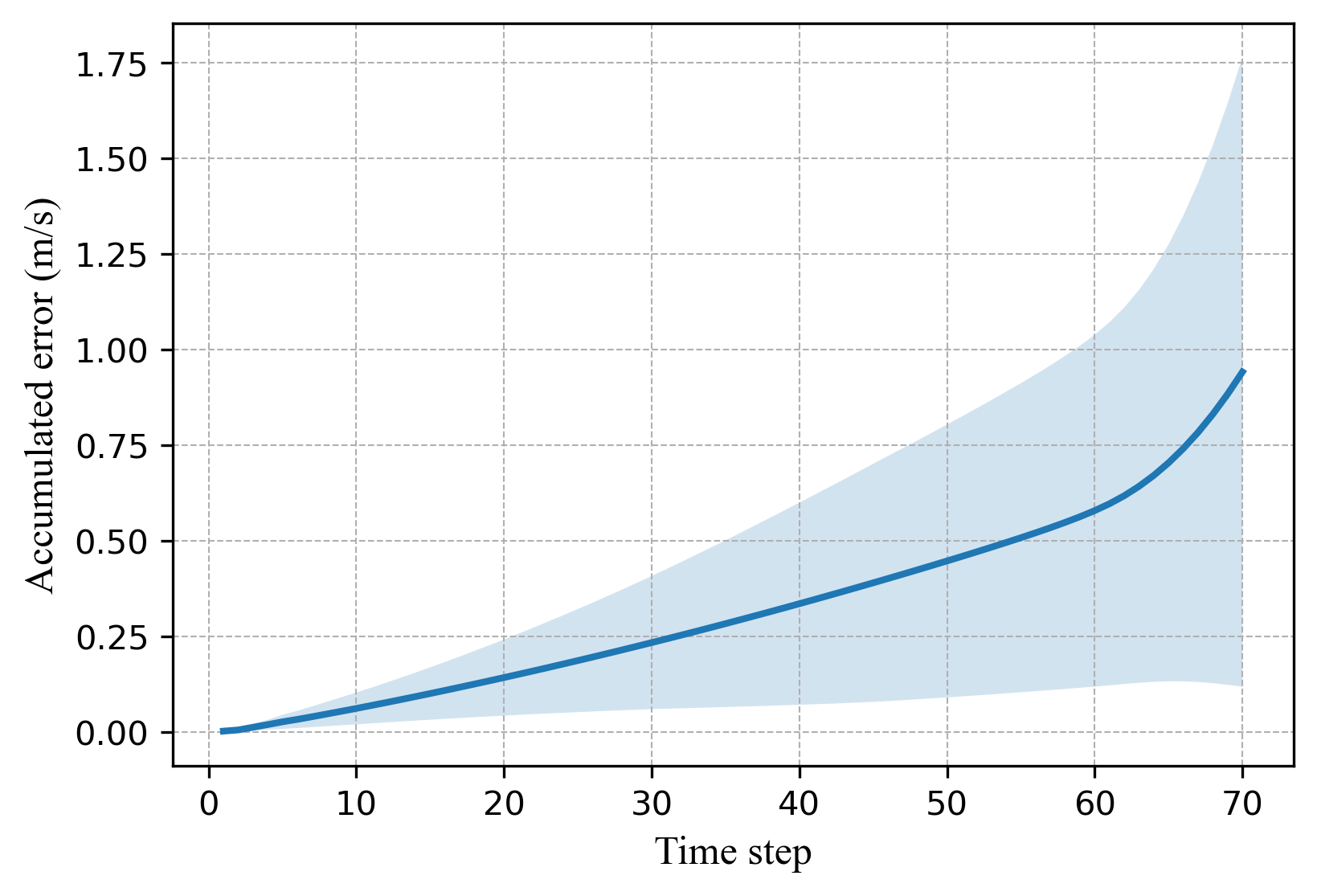}
\caption{Accumulated error of sequential time step prediction}
\label{fig:error_curve}
\end{figure}

We present the detail of the accumulated error of 70 sequential steps in Figure \ref{fig:error_curve}. The accumulated error is computed by taking the mean of the absolute errors across all grids. The colored area in the figure represents the standard deviation of these absolute errors. Numerical experiments show that the accumulated absolute error reaches 0.1 m/s after about 15 steps, and it reaches 1 m/s within 70 steps. During the first 60 time steps, the accumulated error exhibits an approximately linear increase. However, beyond the 60th time step, the accumulated error follows a nearly exponential growth pattern.


\subsection{Computational efficiency comparison}

To compare the computational efficiency of the numerical method with the FNO model, we conduct
simulations using both approaches on the same machine. The evaluations are performed on a Tesla A100 GPU with 40 GB graphical memory and Dual AMD Rome 7742 CPUs. We record the time consumption (in seconds) for computing one time step forward.
Details for generating the simulation data with numerical solver are provided in section data description section. It is worth noting that the numerical solver was specially optimized for GPU usage\cite{mortezazadeh2019cityffd}. Table \ref{tab:time} shows that the FNO approach provides over 25 times speed up compared with the traditional numerical solver, allowing real-time simulations for 3D dynamic urban microclimate.



\begin{table}[ht]
\centering
\caption{Computational efficiency performance}
\label{tab:time}
\begin{tabular}{lll}
\hline
Method           & Scale  & GPU time (s) \\ \hline
\textcolor{black}{CityFFD} & \textcolor{black}{24,000,000 mesh elements}  & 2.21          \\
FNO              & 6,560,581 parameters    & 0.08         \\ \hline
\end{tabular}
\end{table}

\section{Conclusion}

\textcolor{black}{In this study, we first extend the FNO approach for high-resolution fluid dynamics simulation at an urban scale. Prior 3D applications of FNO were limited to a maximum grid size of 64 × 64 × 64 \cite{peng2023linear,li2022fourier}. When dealing with urban-scale simulations, the size of 3D wind field data can easily surpass the memory capacity of a GPU, making it impossible to take advantage of GPU computational speed.}

\textcolor{black}{To address the bottleneck of GPU memory shortage when processing high-resolution 3D data, we propose employing downsampling techniques with cubic spline interpolation. This approach enables us to reduce the 3D wind field data grids from 400 × 400 × 150 to 200 × 200 × 150 while retaining the essential details, thereby making it possible to conduct training on a Tesla A100 GPU. In the realm of FNO applications, our work represents one of the pioneering efforts to adapt FNO to such extensive data size. Moreover, this downsampling approach has the potential to benefit other scientific computing tasks that involve handling high-resolution data.}

Our study demonstrates that the FNO approach has the capacity to significantly expedite the simulation of the 3D urban wind field while maintaining a high level of accuracy in reconstructing the instantaneous spatial velocity magnitude field, with an error rate of less than 1\%. \textcolor{black}{Furthermore, we evaluate the trained FNO model on unseen data with different wind directions. Results show that the FNO model, trained solely on data of one wind direction, can generalize well to distinct wind directions. Although previous studies have demonstrated the benefits of FNO in generalizing to various initial conditions when solving Navier-Stocks equations \cite{li2020fourier, guibas2021adaptive, li2023long}, our research stands out as the first to validate these advantages in the context of urban-scale microclimate simulation with diverse wind directions. It opens new possibilities for high-quality AI-accelerated urban microclimate simulations within intricate environmental settings.}

\section{Discussion and future work}

While FNO offers advantages in simulating urban microclimates at high resolutions, there are several limitations that demand further exploration in future research.




The first limitation is that the current FNO model can only predict the absolute wind speed, due to the large size of high dimensional data, and constrained GPU memory.  In future work, we aim to develop more efficient and lightweight models and predict more variables including the velocity components and temperature field.

The second limitation is that the FNO model is trained at fixed wind speed and fixed direction. Despite that its generalization ability has been tested on varying wind directions, urban microclimate simulations in real-world applications are even more complex and challenging. In future work, we plan to improve the FNO model generalization performance on different wind speeds and different cities. \textcolor{black}{The training data we generate from the numerical solver is under a constant wind speed and direction. Due to the chaotic nature of wind fields in the real world, it's important to investigate the performance of the FNO model under wind fields with dynamic speed and direction. }

The third limitation is that the error increases with time due to the chaotic features of the dynamic system. Therefore the prediction errors are produced and accumulated at
each step. In future work, we intend to reduce the temporal prediction error using approaches such as the attention mechanism \cite{peng2022attention,peng2023linear,liu2022graph,deo2022learning,kissas2022learning}, or incorporating the physical constraints to ensure that the predictions are subject to the governing equations and conservation laws \cite{raissi2019physics,kashinath2020enforcing,li2021physics,kashefi2022physics,costa2023deep}. Furthermore, by utilizing real-time observations, the integration of machine learning with data assimilation \cite{geer2021learning, buizza2022data, bocquet2020bayesian, brajard2020combining} shows great promise for enhancing the simulation of complex and possibly chaotic dynamics, leading to both improved efficiency and reduced accumulated error.

\textcolor{black}{In future work we also plan to include more morphological parameters as input during training. Information like the size and density of the buildings, along with the spatial dimensions between the buildings, enables the FNO model to acquire a more comprehensive understanding of the physical relation intrinsic to the environment and thus improve generalization capacity.}

\section*{Data availability statement}

The data and code that support the findings of this study are available
from the corresponding author upon reasonable request.


\bibliography{main}






\end{document}